\documentclass[conference]{IEEEtran}

\usepackage{cite}
\usepackage{amsmath,amssymb,amsfonts}
\usepackage{algorithmic}
\usepackage{graphicx}
\usepackage{textcomp}
\usepackage{xcolor}
\usepackage{subfigure}

\usepackage{stackengine}
\usepackage{tabularray}

\def\BibTeX{{\rm B\kern-.05em{\sc i\kern-.025em b}\kern-.08em
    T\kern-.1667em\lower.7ex\hbox{E}\kern-.125emX}}

\begin{document}

\title{A Hierarchical Heuristic for Clustered Steiner Trees in the Plane with Obstacles}

\author{\IEEEauthorblockN{Victor Parque}
\IEEEauthorblockA{\textit{Graduate School of Advanced Science and Engineering} \\
\textit{Hiroshima University}\\
1-4-1 Kagamiyama, Higashi-Hiroshima, Hiroshima Prefecture, 739-8527, Japan \\
parque@hiroshima-u.ac.jp, https://orcid.org/0000-0001-7329-1468}
}

\maketitle

\begin{abstract}

Euclidean Steiner trees are relevant to model minimal networks in real-world applications ubiquitously. In this paper, we study the feasibility of a hierarchical approach embedded with bundling operations to compute multiple and mutually disjoint Euclidean Steiner trees that avoid clutter and overlapping with obstacles in the plane, which is significant to model the decentralized and the multipoint coordination of agents in constrained 2D domains. Our computational experiments using arbitrary obstacle configuration with convex and non-convex geometries show the feasibility and the attractive performance when computing multiple obstacle-avoiding Steiner trees in the plane. Our results offer the mechanisms to elucidate new operators for obstacle-avoiding Steiner trees.

\end{abstract}


\section{Introduction}

Minimal-length trees enable to model optimal networks in the plane. Since the inquiry by Fermat on how to minimally connect three points in the plane, and the formalization by Gilbert-Pollack\cite{gilbert68}, minimal Steiner trees has received relevant attention due to applications in graph query, network design, network location, VLSI, biology, mathematical modeling, supply chain management, and multi-agent planning problems. 

The Euclidean Steiner Tree problem (EST) aims at interconnecting $n$ terminal nodes by a shortest possible network while allowing the addition of extra nodes whenever necessary (the extra nodes are also known as the Steiner points or Fermat points)\cite{geo18}. When obstacles in the plane are involved, the edges of the minimal Steiner tree are to avoid overlapping with the obstacles. The community has studied polynomial algorithms and approximation heuristics for tailored ESTs\cite{brazil14}. For instance, Winter and Smith proposed in the 90s an \(O(n)\) approach considering three terminals and one obstacle \cite{winter91}. Later on, \cite{winter93} used visibility graphs to join ESTs with up to four terminals. Zachariasen and Winter concatenated full Steiner trees\cite{zacha99}. Weng and Smith proposed the polynomial algorithm considering extreme points in obstacles with convex geometry \cite{weng01}. Winter et al. concatenated ESTs with four terminals and considered a simple polygon\cite{winter02}. In 2010, Müller-Hannemann and Tazari proposed a polynomial-time approximation with \(O\left(n \log ^{2} n\right)\), for \(n\) being the number of terminals and obstacle vertices \cite{muller10}. Cohen and Nutov studied enhanced approximation ratios for Steiner trees and forests \cite{cohen18}. Chen et al. tackled the obstacle-avoiding quadrilateral Steiner trees with application to wireless networks \cite{chen19}. Brazil et. al. \cite{brazil24} proposed topological pruning methods for ESTs using at most $k$ additional Steiner points, and \cite{volz22} simplified obstacle configuration for obstacle-avoiding ESTs.

The community has proposed gradient-free approaches to tackle EST problems. For instance, the slime mold networks\cite{caleffi15} and physarum-inspired algorithms\cite{sun16} were competitive against Genetic Algorithms (GA) and Particle Swarm Optimization (PSO). Differential Evolution (DE) effectively aided to construct obstacle-avoiding ESTs\cite{ictai18,zavo18}, and the dividing rectangles (DIRECT) improved the convergence\cite{evocop21}, the bi-level and coevolution formulations were found to be practical\cite{camacho19}, whereas Variable Neighborhood Search (VNS)\cite{chuo19} and (1+ 1) Evolutionary Strategy\cite{lai17} were effective for Steiner trees in graphs. Furthermore, minimal trees have been studied in circuit design\cite{vlsi20} (VLSI systems) and pipe routing problems\cite{case18} (offshore oil and gas industries). In particular, in octilinear EST (also referred to as X-architecture), edges in the minimal network are oriented at \(0^{\circ}, 45^{\circ}, 90^{\circ}, 135^{\circ}\). As such, the community has used Particle Swarm Optimization\cite{huang15}, Genetic Algorithms\cite{liu18}, Ant Colony Optimization \cite{pipe15}.

The generation of multi-Steiner trees has recently attracted the attention of the community. For instance, in a recent study, \cite{long24} tackled the clustered Steiner tree problem using a multifactorial evolutionary optimization on weighted undirected graphs, \cite{meng23} studied the partitioned Steiner tree problem on undirected graphs based on branch-and-cut and integer linear programming, and \cite{cmu22} studied the hardness and the approximation factors for Hub Steiner trees in graphs. Although the existing approaches use nature-inspired and concatenation algorithms to construct single ESTs that avoid overlap with obstacles, the study of efficient schemes to generate multi-ESTs which are free of clutter and overlapping with arbitrary obstacle geometry has received little attention in the community. As such, in this paper, we study an approach based on hierarchical bundling schemes to compute multiple obstacle-avoiding Euclidean Steiner trees in the plane. In particular, our contributions are as follows:

\begin{figure*}[h!]
\centering
\hfill
\subfigure[Obstacles and Nodes]{\includegraphics[width=0.21\textwidth]{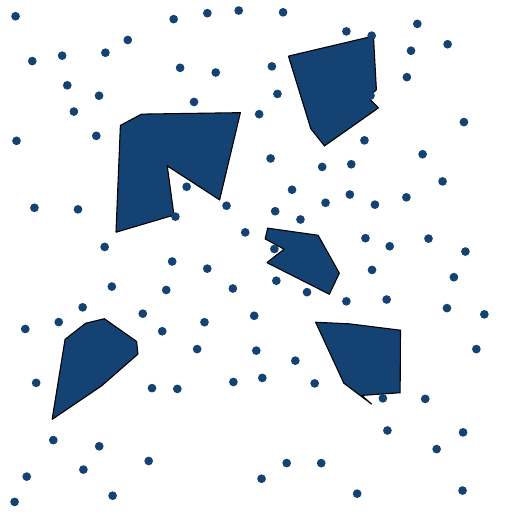}}
\hfill
\subfigure[5 modules of Steiner trees]{\includegraphics[width=0.24\textwidth]{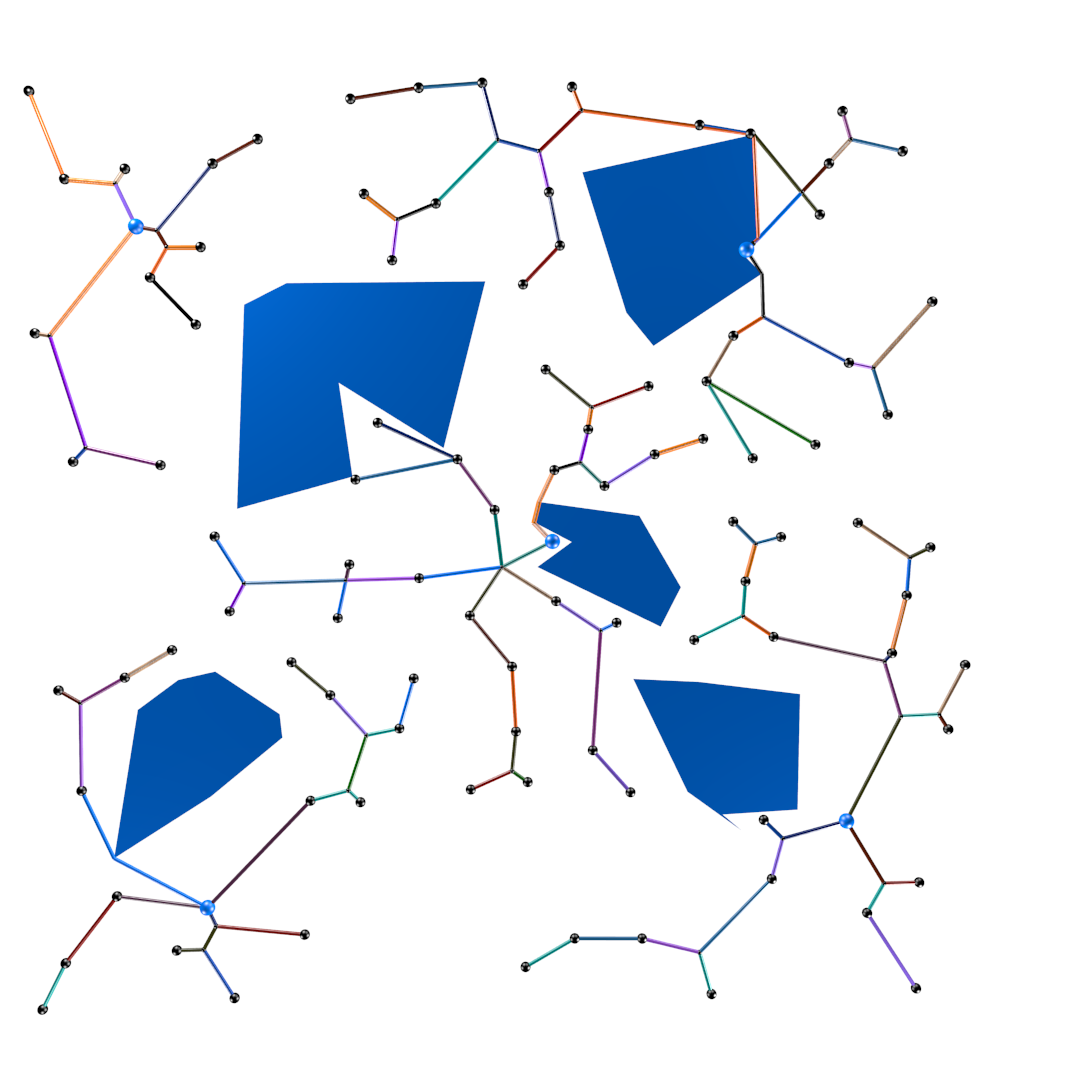}}
\hfill
\subfigure[15 modules of Steiner trees]{\includegraphics[width=0.24\textwidth]{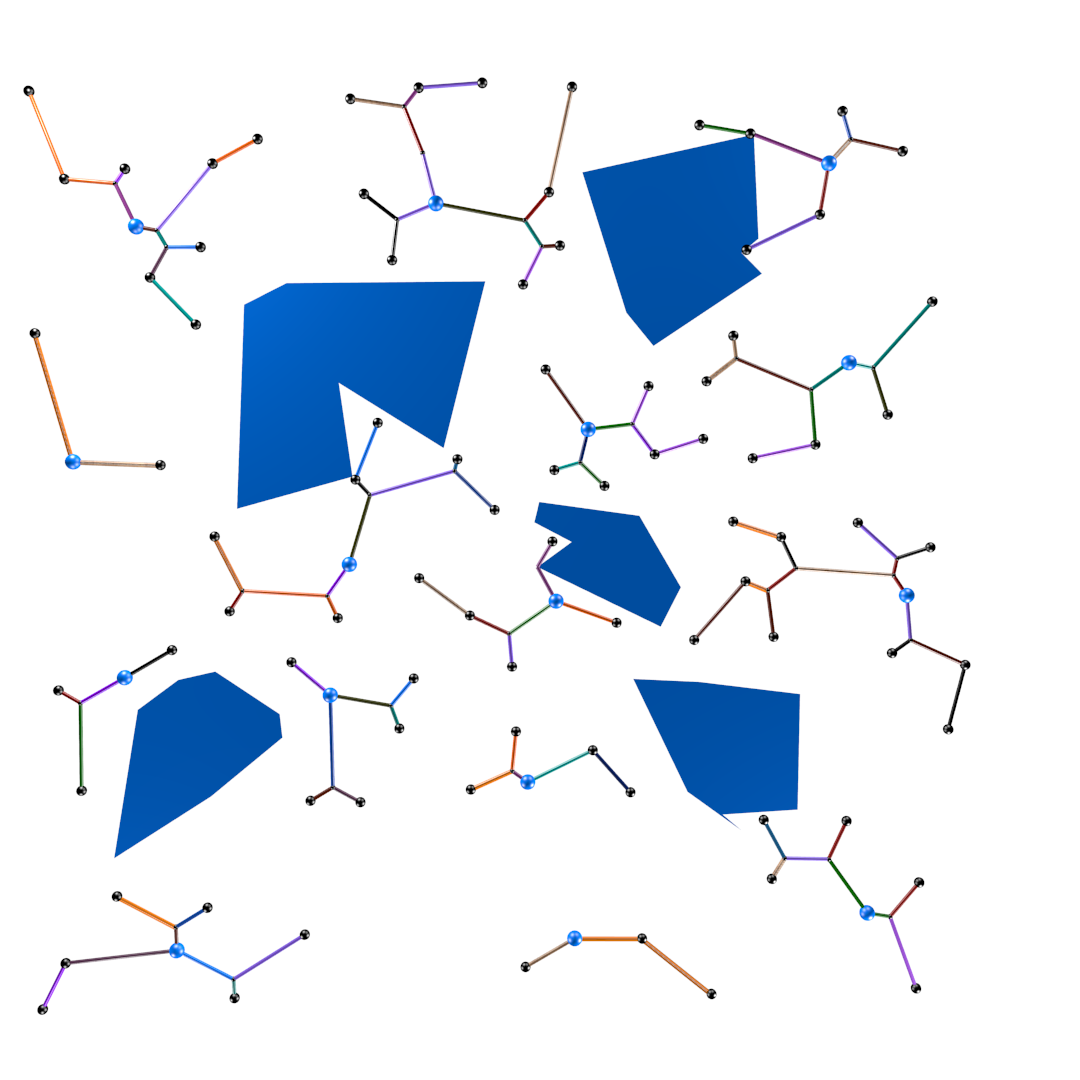}}
\hfill
\subfigure[25 modules of Steiner trees]{\includegraphics[width=0.24\textwidth]{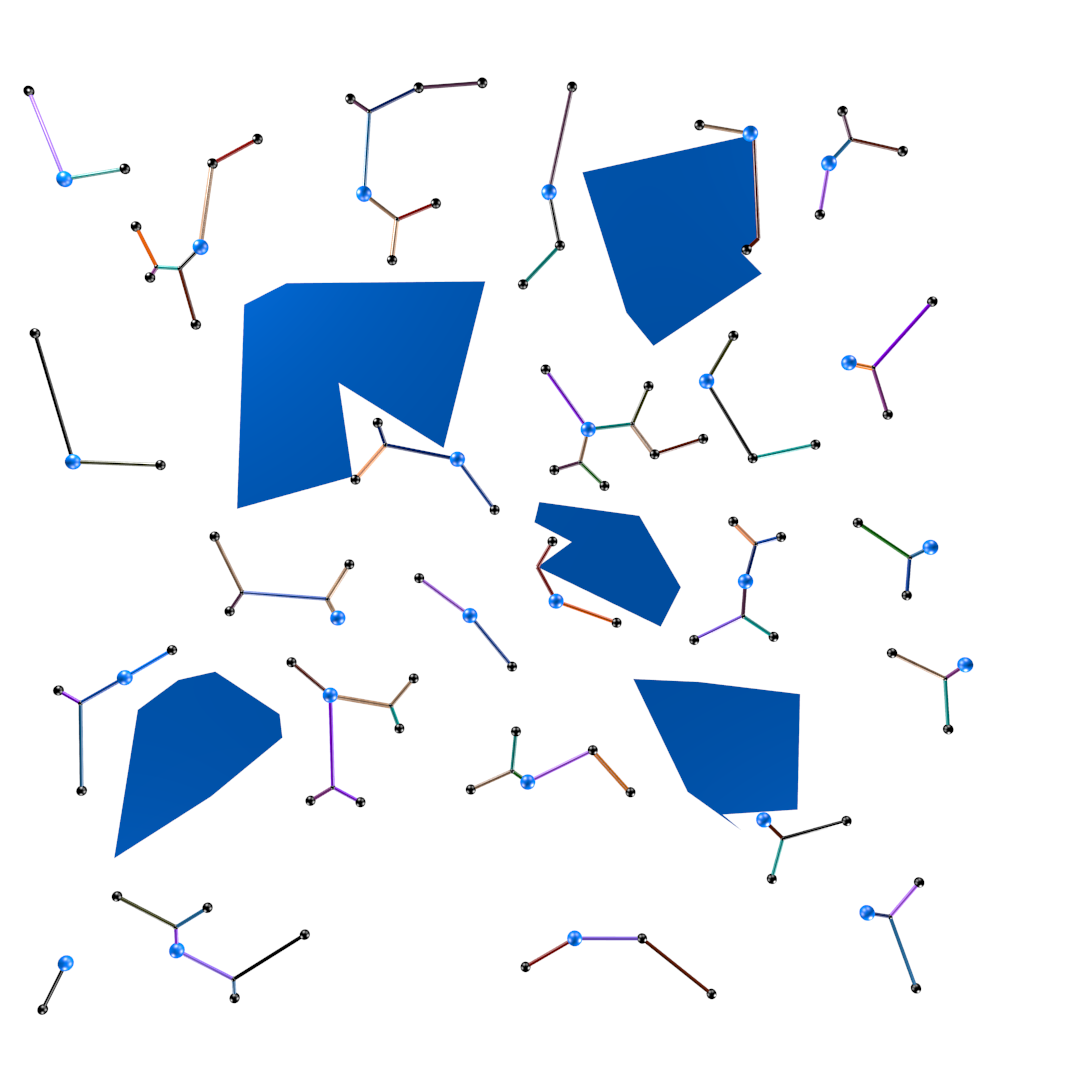}}
\hfill
\caption{Example of (a) the configuration of a plane with obstacles with either convex and non-convex geometries and terminal nodes located in the free space, (b)-(d) obstacle-avoiding multi-Euclidean Steiner trees on the plane with 5, 15, and 25 modules.}
\label{ex}
\end{figure*}

\begin{itemize}
  \item We propose a hierarchical approach for clustered Steiner trees in the plane. Our method uses hierarchical trees to construct and concatenate multi-Euclidean Steiner trees via hierarchical bundling of routes and trees, effectively avoiding clutter and overlapping with arbitrary obstacles.
  \item Our computational experiments using arbitrary maps with convex and non-convex obstacle configurations show the merit and the feasibility of computing multiple and mutually disjoint Steiner trees on the plane.
\end{itemize}

\section{Obstacle-Avoiding Multi-Euclidean Steiner Trees}
In this section we describe the overall framework and basic ideas behind our proposed approach.

\subsection{Preliminaries}
We aim at constructing multiple Steiner trees that connect a set of points in the plane while avoiding clutter and overlap with obstacles. As such, we consider the following inputs as given a-priori:

\begin{itemize}
  \item A set of terminal nodes \(P=\left\{P_{1}, P_{2}, \ldots, P_{n}\right\} \) whose locations are given in the 2-dimensional Euclidean plane, and
  \item A set of obstacles \(O=\left\{O_{1}, O_{2}, \ldots, O_{\lambda}\right\}\) in the Euclidean plane represented by polygons of arbitrary geometry (in this paper, for simplicity and without loss of generality, we consider obstacles with arbitrary geometry, e.g. convex and non-convex geometries).
\end{itemize}

To exemplify the above-mentioned, Fig. \ref{ex}-(a) shows the configuration of a set of terminal nodes (as points/dots in the 2-dimensional domain) and a set of polygons (obstacles) with arbitrary geometry. Here, the number of terminal nodes is \(n=100\), and the number of obstacles is \(\lambda=5\). The question we address in this paper is how to generate multiple Euclidean Steiner trees that connect \(n=100\) nodes while avoiding clutter and overlapping with both trees and obstacles. Fig. \ref{ex} (b)-(d) show the configuration of multiple Steiner trees associated to the input set in Fig. \ref{ex}-(a). Here, for clarity of the explanation, we portray the cases when generating \(\{5, 15, 25\}\) modules of Steiner trees. By observing Fig. \ref{ex} (b)-(d), we note that minimal trees connect the terminal nodes by using modules (groups), and that minimal trees are able to avoid clutter and overlap with obstacles.

The key motivation behind the idea of constructing multiple Euclidean Steiner trees in the plane is inspired by the need of considering the deployment of teams of multi-agents (robots, sensors, hubs) in the environment, thus the configuration of Steiner trees allows to model not only efficient information/communication/distribution networks, but also the decentralized/modular control schemes. In this application domain, the obstacle geometry is of convex or non-convex nature.

The basic idea to generate multiple Steiner trees is as follows:

\begin{enumerate}

  \item Cluster the set of terminal nodes \(P\) by a hierarchical (agglomerative) clustering scheme, rendering the cluster tree $Z$ (dendrogram). And for a given depth of $Z$, find $s$ clusters and terminal subset \(M=\left\{M_{1}, M_{2}, \ldots, M_{i}, \ldots, M_{s}\right\}\), $M \subset P$.
      
  \item For each cluster in the set $M$, generate a Steiner tree using the hierarchical bundling of shortest paths.
      
  \item To generate the more compact structures of Steiner trees, concatenate (bundle) Steiner trees by using the tree $Z$ and by repeating step (2) until a user-defined criterion is met.
  
\end{enumerate}

\subsection{Hierarchical Clustering}

To generate the cluster tree $Z$, we used the hierarchical clustering scheme with complete linkage (farthest distance) and the Euclidean norm between two coordinates as a distance metric. Also, for a given depth of $Z$, we find $s$ clusters by finding the smallest depth in the cluster tree $Z$ that renders $s$ or fewer modules; in which the set of modules (clusters) is labeled with \(M=\left\{M_{1}, M_{2}, \ldots, M_{i}, \ldots, M_{s}\right\}\), and cluster \(M_{i} \subset P\) for \(i \in[1, s]\). The set of nodes in the cluster $M_i$ becomes non-intermediate nodes of the Steiner trees.

\subsection{Hierarchical Bundling of Shortest Paths}

\begin{figure*}[t]
\centering
\includegraphics[width=0.98\linewidth]{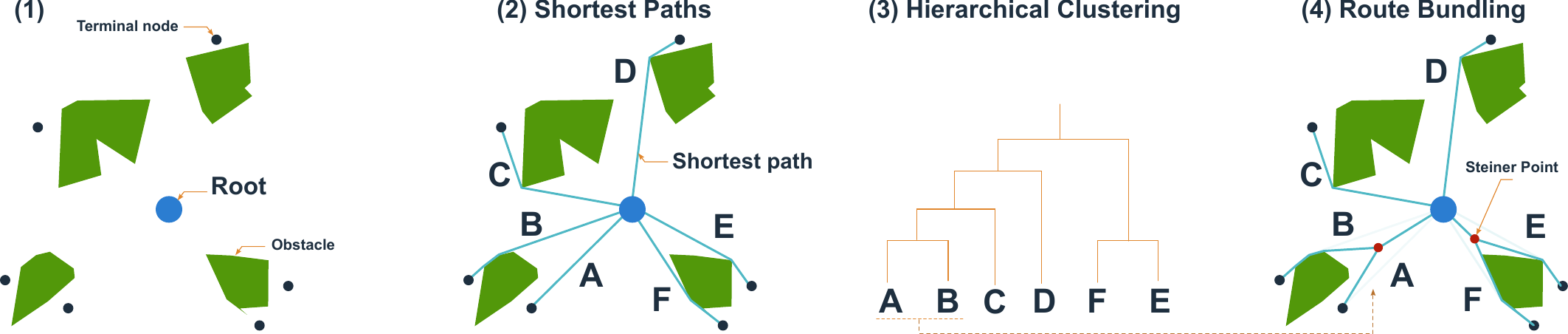}
\caption{Basic idea of a hierarchical bundling of shortest paths.}
\label{bundling}
\end{figure*}

To generate multiple Steiner trees, we find the root node \(R_{i}\) in cluster $M_i$ with minimal distance to other nodes. Due to the above, the root \(R_{i} \in M_{i}\) and \(R_{i} \in P\). And for each cluster \(M_{i}\), we construct a Steiner tree $t_i$ by using a hierarchical bundling of the set of shortest paths from root \(R_{i}\) to every terminal node in $M_i$, whose basic idea is depicted by Fig. \ref{bundling}. Here, we assume a cluster with proximal terminal nodes being identified on the plane as shown in Fig. \ref{bundling}-(1). Let a cluster be defined with

\begin{equation}\label{eqM}
  M_{i}=\left\{P_{1}^{i}, P_{2}^{i}, \ldots, P_{k}^{i}, \ldots, P_{m_i}^{i}\right\} \cup\left\{R_{i}\right\}
\end{equation}

where \(P_{k}^{i}\) is the \(k\)-th terminal node in the \(i\)-th cluster, $m_i$ in the number of terminal nodes in the $i$-th cluster, and \(R_{i}\) is the root of the \(i\)-th cluster. Then, we construct a Steiner tree $t_i$ by the following procedure:

\begin{figure*}[t]
\centering
\includegraphics[width=0.98\linewidth]{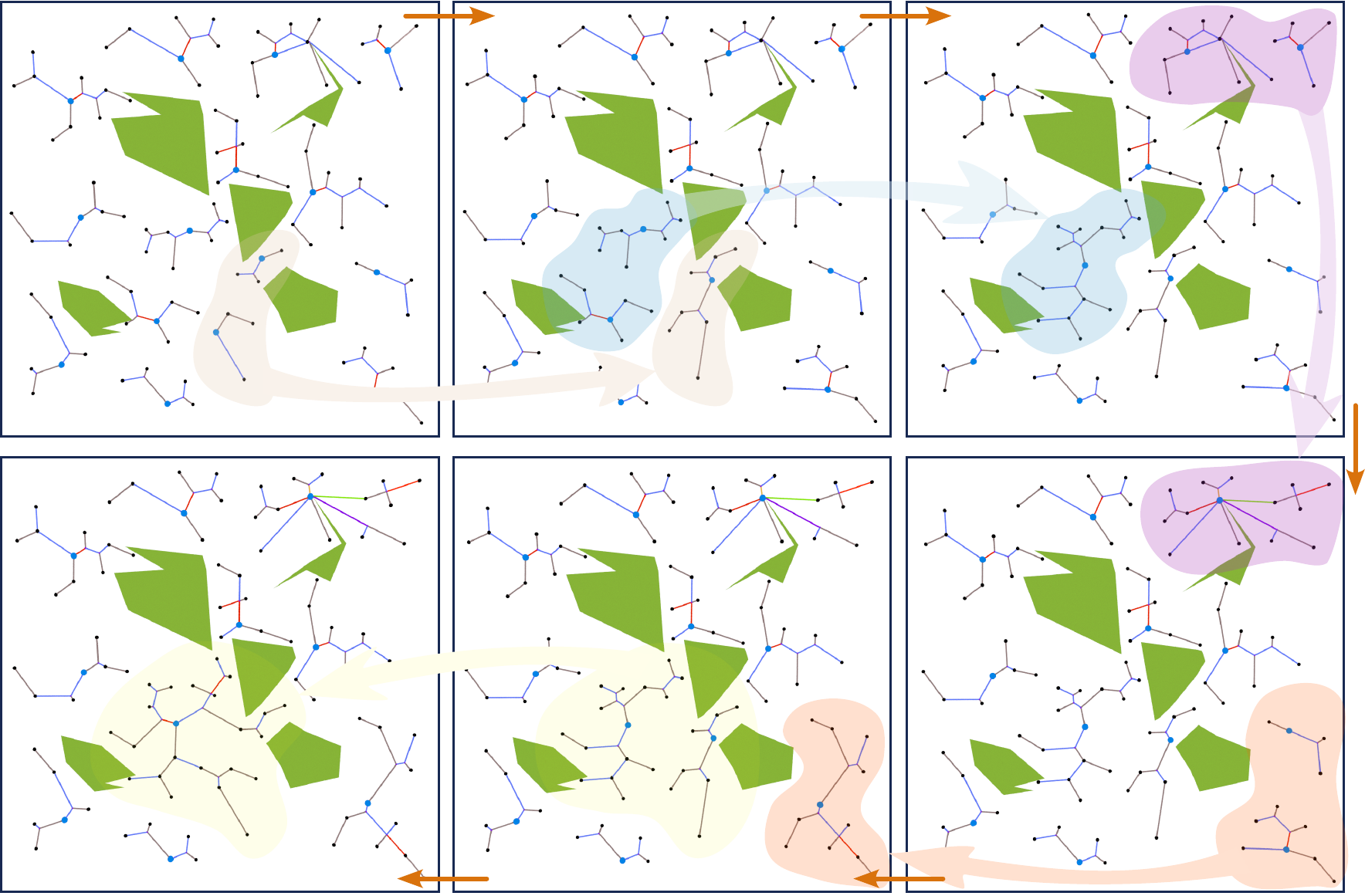}
\caption{Examples of the concatenation of multi-Steiner trees in the plane with obstacles. The colored regions indicate areas where concatenation of Steiner trees occur, following the sequence of arrows in orange color.}
\label{concat}
\end{figure*}

\begin{enumerate}

  \item First, for each cluster \(M_{i}, ~i \in[1, s]\), we compute the set of shortest paths from the root \(R_{i}\) to every other terminal node in \(M_{i}-\left\{R_{i}\right\}, i \in[1, s]\). The set of shortest paths is represented by the set of tuples

  \begin{equation}\label{eqS}
    S_i = \{ (R_i, ..., P^i_k), ~ i\in [1, s], k \in [1, m_i] \}, 
  \end{equation}
  
  To exemplify the above, Fig. \ref{bundling}-(2) shows the root \(R_{i}\) in blue-colored sphere and the shortest paths are depicted with dark-colored lines and labeled with symbols \textsf{A, B, C, D, E, F}. To compute the shortest paths in arbitrary convex and non-convex polygonal domains, we use the A* algorithm with visibility graphs.
      
  \item Second, we cluster the set of shortest paths \(S_{i}\) by a hierarchical agglomerative approach with a distance metric as follows:

\begin{equation}\label{dpath}
  d\left(S_{i, u}, S_{i, v}  \right) = G(u, v) \cdot  \cos ^{-1}\left(\frac{\boldsymbol{a} \cdot \boldsymbol{b}}{|\boldsymbol{a}||\boldsymbol{b}|}\right) 
\end{equation}

\begin{equation}\label{eqG}
  G(u, v)=\displaystyle \sum_{h=1}^{NP}\left\|S_{i,u}^{h} - S_{i,v}^{h}\right\|^{2} 
\end{equation}

\begin{equation}\label{eqa}
  \boldsymbol{a}=P_{k_u}^{i}-R_{i} 
\end{equation}

\begin{equation}\label{eqb}
  \boldsymbol{b}=P_{k_v}^{i}-R_{i}
\end{equation}
where $d(S_{i, u}, S_{i, v})$ denotes the similarity between the $u$-th and $v$-th shortest paths from the set $S_i$, $S_{i,u}^{h}$ is the \(h\)-th coordinate sampled from the shortest path $S_{i,u}$, $NP$ is the number of points for piecewise interpolation between shortest paths, the vector \(\boldsymbol{a}\) describes the orientation from the root \(R_{i}\) to terminal node $P_{k_u}$ of the $u$-th shortest path, and $ ||.||$ denotes the Euclidean norm. The result of the hierarchical clustering of shortest paths is a dendrogram \(Z_i\) as shown in Fig. \ref{bundling}-(3) which encodes the similarity structure among shortest paths and modules hierarchically.

The distance metric used in Eq. \ref{dpath} has two roles: to measure the piecewise gaps $G(u, v)$ and to estimate the orientation gaps to the terminal nodes through the difference of angles. As such, piecewise and orientation gaps are considered to construct the hierarchy $Z_i$. This choice implies that overlapped shortest paths sharing similar directions are considered close, thus they are subject to bundling operations through the ordering of the dendrogram $Z_i$. Exploring distinct distance metrics has the potential to explore tailored forms of Steiner tree generation.

\item Third, we compute Steiner points by searching for intermediate points lying in the convex hull of the shortest paths rendered from the dendrogram \({Z_i}\). The goal is to find Steiner points that minimize tree length. The search for intermediate points is realized through trust region algorithm based on Sequential Quadratic Programming (SQP)\cite{nelder}. To enable the efficient search over the plane with obstacles, we sample solutions over the triangulated feasible search space. In particular, the representation of Steiner points is as follows

    \begin{equation}\label{eqr}
      r=\left(\tau, r_{1}, r_{2}\right)
    \end{equation}
    
    where \(\tau\) is an integer number encoding the order of the triangle derived from the Delaunay triangulation of the free space, and \(r_{1}, r_{2} \in[0,1]\). The above representation maps to the Cartesian coordinate\cite{osada02}:

    \begin{equation}\label{rcart}
      \left(r_{x}, r_{y}\right)=\left(1-r_{1}\right) a_{\tau}+\sqrt{r_{1}}\left(1-r_{2}\right) b_{\tau}+\sqrt{r_{1}} r_{2} c_{\tau}
    \end{equation}

    where \(a_{\tau}, b_{\tau}, c_{\tau}\) are the Euclidean coordinates of the vertices of the \(\tau\)-th triangle derived from the triangulated search space (Delaunay triangulation from the visibility graph of the map). The above-mentioned representation scheme is advantageous to enable the efficient search and sampling of feasible Steiner points that implicitly avoid overlap with obstacles, rendering a convex search space in which checks of point inside obstacles becomes unnecessary.

    Furthermore, the bundling of shortest paths is performed through a recursive/hierarchical approach guided by the tree structure $Z_i$. Thus, to consider the local and global topology of the tree, the bundling operators are performed in bottom-up (from terminal to root) and top-down (from root to terminal nodes).

\end{enumerate}

\begin{figure*}[t]
\centering
\includegraphics[width=0.98\linewidth]{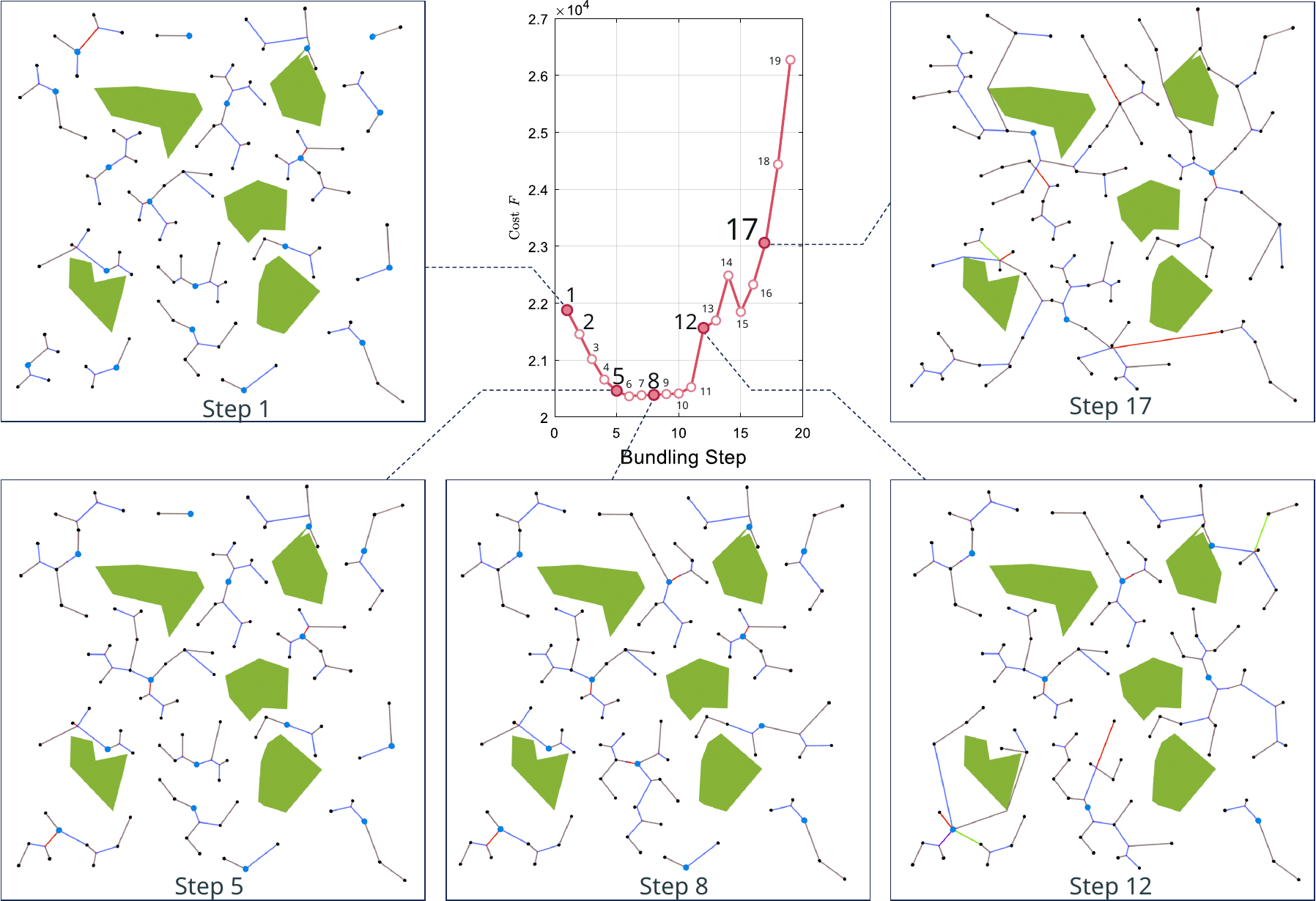}
\caption{Examples of the landscape of a cost function and the corresponding configuration multi-Steiner trees in the plane.}
\label{bcostmap}
\end{figure*}

\begin{figure*}[t]
\centering
\includegraphics[width=0.98\linewidth]{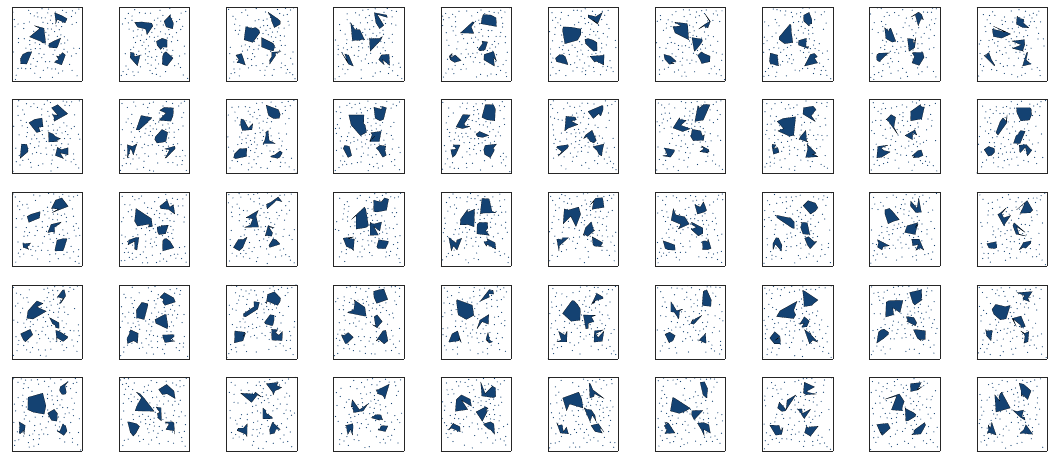}
\caption{The rendering of 50 arbitrary configurations of terminal nodes and obstacles in the plane.}
\label{maps50}
\end{figure*}

\subsection{Concatenation of Steiner Trees}

To generate the more compact and larger Steiner tree structures, we construct new Steiner trees by concatenating (bundling) Steiner trees and by using the hierarchical tree $Z$ computed in step 1 of section II-A. 

Since points clustered into the cluster tree $Z$ encodes both the closeness and the similarity between Steiner trees, the cluster $Z$ is able to guide the construction of more compact Steiner trees by concatenation/bundling operations. As such, to concatenate Steiner trees $t_i$ and $t_j$, each of which is associated to modules $M_i$ and $M_j$, respectively, we find the joint set of terminal nodes $M_{i,j} = M_i \cup M_j$, and perform the hierarchical bundling of shortest paths over the set $M_{i,j}$ (section II-C). 

\begin{equation}\label{eqbun}
  t_{i,j} = t_i \oplus t_j
\end{equation}
where $t_i$ is the $i$-th Steiner tree rendered from set $M_i$, $\oplus$ denotes a bundling operator between Steiner trees, and $t_{i,j}$ denotes the concatenated (bundled) Steiner tree. Bundling Steiner trees implies updating the topology of the Steiner tree as a result of updating the root and Steiner points. For efficiency considerations, the concatenation (bundling) of Steiner trees brings the following benefits:

\begin{itemize}
  \item It becomes possible to re-use the computed shortest paths from the component sets $S_i$ and $S_j$, since they share common elements with the set $S_{i, j}$. 
  \item Since computing the Steiner tree points occurs within the local search space of the convex hull of closest shortest paths, it becomes possible to re-use the previously computed Steiner Points as the initial guess for subsequent Sequential Quadratic Programming (SQP) optimization in step (3) of section II-C.
  \item Furthermore, it becomes possible to re-use computed distances from the roots of component Steiner trees to terminal nodes.
\end{itemize}

Since the concatenation of Steiner trees is guided by the binary cluster tree $Z$, the concatenation of Steiner trees is a binary bottom-up process that aims at joining Steiner tree components into larger structures. Fig. \ref{concat} renders an example of the concatenation of Steiner trees following a binary bottom-up approach (depicted by the corresponding colored regions and following the sequence of orange-colored arrows). By observing Fig. \ref{concat}, one can observe the feasibility of computing larger tree structures and the corresponding changing topologies and root locations. Also, one can observe from Fig. \ref{concat} that the concatenation/bundling of Steiner trees follows a sequential process that is based on the cluster tree $Z$ and is expected to end when a user-defined criterion is met. In this paper, we propose using the following criterion:

\begin{equation}\label{equser}
F = w_l L_t + w_d L_d
\end{equation}

\begin{equation}\label{Lt}
L_t = \sum_{i = 1}^{s} L(t_i)
\end{equation}

\begin{equation}\label{Ld}
L_d = \sum_{1 \leq i < j \leq s } ||R_i - R_j||
\end{equation}
where $L_t$ is the sum of the lengths of all the Steiner trees in the plane, $L_d$ is the sum of the distances among the roots of the Steiner trees, and $w_l$ and $w_d$ are user-defined coefficients. Whereas $L_t$ aims at measuring the size of Steiner trees, $L_d$ aims at measuring the separation between roots, as potential hubs, corresponding to the entire set of Steiner trees. Thus, finding the optimal configuration of multi-Steiner trees implies minimizing Eq. (\ref{equser}) for user-defined coefficients $w_l$ and $w_d$. Given that Steiner tree concatenation follows a sequential process guided by the cluster $Z$, Eq. (\ref{equser}) naturally serves as a guiding criterion to evaluate the overall Steiner tree bundling process. 

To exemplify such mechanism, Fig. \ref{bcostmap} denotes the landscape of a cost function that corresponds to Eq. (\ref{equser}) for fixed $w_l$ and $w_d$. In Fig. \ref{bcostmap}, the x-axis denotes the order of the concatenation (bundling) operation, and the y-axis denotes the value of the cost function computed by Eq. (\ref{equser}). The cost function of the initial set of Steiner trees can be measured when the bundling step is equivalent to one (Step 1 as shown in Fig. \ref{bcostmap}). Then, subsequent concatenation (bundling) operations render multi-Steiner tree configurations with a lower cost function. However, the landscape of the cost function reaches a flat region starting at bundling step 5-6 (as shown by Step 6 and Step 8 in Fig. \ref{bcostmap}). Such flat regions in the landscape of the cost function suggest the desirable circumstances to stop the concatenation (bundling) of Steiner trees. For the purpose of the example, Fig. \ref{bcostmap} also shows the topologies for Step 12 and Step 17, both of which render an elevated cost function. The landscape of Fig. \ref{bcostmap} is dependent upon the configuration of nodes and obstacles in map, and the setting of user-defined coefficients $w_l$ and $w_d$. In a later section, we describe the potential configuration for a larger number of scenarios and portray observations on the performance landscape over a relevant number of scenarios.

\section{Computational Experiments}

In order to evaluate the feasibility and the effectiveness of our proposed approach, we performed computational experiments portraying the generation of multiple Steiner trees on the Euclidean plane. This section describes our experimental settings and obtained observations.

\subsection{Settings}

To evaluate the feasibility of generating multi-Steiner trees that avoid clutter and overlap with arbitrary obstacles in the environment, we have generated arbitrary geometries of obstacles and configurations of terminal nodes in the plane. Thus, for our environments for evaluation, we used the following setup:

\begin{itemize}
  \item We used 50 maps in the 2-dimensional domain bounded by a square of 200 by 200 units. Although we use A* algorithm for shortest path finding, our approach is independent of the size of the map due to the fact of computing visibility graphs. 
  \item Each map contained five non-overlapping obstacles modeled by polygons with 7 edges considering both convex and non-convex geometry. The generation of obstacle geometry and location within the map was realized through a randomized scheme.
  \item For each map in the 2-dimensional domain, we arbitrarily generated 100 (terminal) nodes over the free navigable space, that is regions outside the obstacle domains.
  \item Also, for each map in the 2-dimensional domain, we considered the computation of diverse configurations of Steiner tree modules corresponding to a parameter \(\theta\) which represents a fraction of the number of terminal nodes $n$. In particular, we used the ratios \(\theta=\{0.05,0.1,0.15,0.2,0.25,0.3,0.35,0.4,0.45,0.5\}\).
  \item Fig. \ref{maps50} shows the configuration of maps and terminal nodes. Here, obstacles are depicted with blue-colored polygons and terminal nodes are shown with blue-colored dots. 
\end{itemize}

\begin{figure*}[t]
\centering
\includegraphics[width=0.98\textwidth]{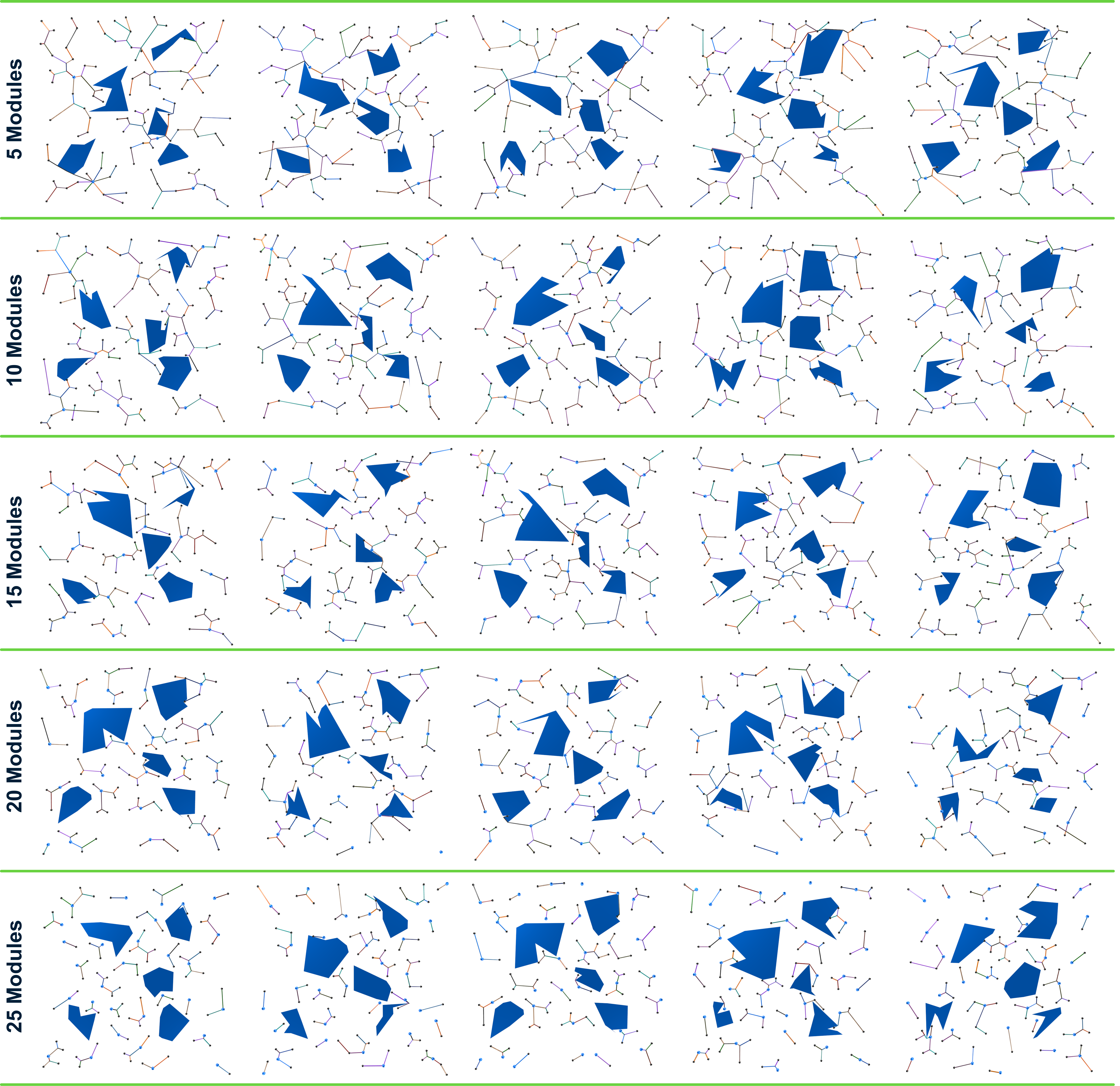}
\caption{Examples of generated multi-Euclidean Steiner trees in the plane with obstacles. Obstacles with arbitrary geometry configuration is shown by blue-colored polygons. The root of each Euclidean Steiner tree is shown in a blue-colored sphere, whereas terminal nodes are shown in black-colored spheres. Edges of the trees are shown with straight lines, each with a distinct color.}
\label{mapex}
\end{figure*}

\begin{figure*}[t]
\centering
\includegraphics[width=0.98\linewidth]{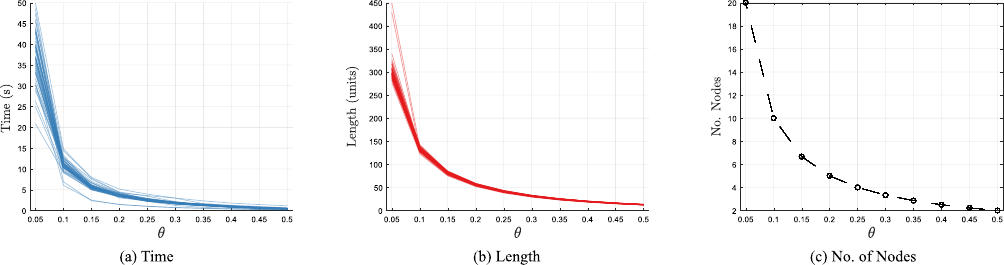}
\caption{Performance in terms of (a) computing time, (b) Steiner tree length and (c) number of nodes per Steiner tree module configuration.}
\label{per50v2}
\end{figure*}

\begin{figure*}[t]
\centering
\includegraphics[width=0.98\linewidth]{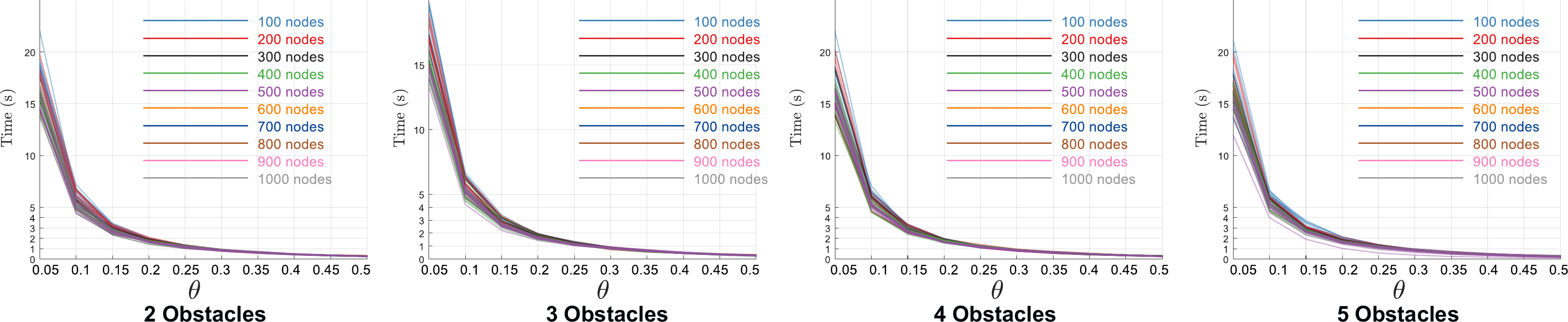}
\caption{Performance in terms of computing time when varying the number of obstacles and nodes in the plane.}
\label{perv3}
\end{figure*}

Due to the above configuration, we generated \(s=\{5,10,15, 20,25,30,35,40,45,50\}\) multiple Steiner trees per map, and rendered 500 configurations of Steiner tree modules in the 2-dimensional domain. Implementations were realized in Matlab 2020a through an Intel i7-4930K @ 3.4 GHz, and hierarchical clustering used the nearest distance as the main agglomeration strategy.

\subsection{Results and Discussion}
In order to show a glimpse of the kind of trees that our proposed approach is able to generate, Fig. \ref{mapex} shows examples of multiple Euclidean Steiner trees on the plane for \(s=\{5,10,15,20,30\}\). By observing Fig. \ref{mapex}, we observe the following facts:

\begin{itemize}
    \item The configuration of obstacles are shown by blue-colored polygons. The root of each Euclidean Steiner tree is shown in a blue-colored sphere, whereas terminal nodes are shown in black-colored spheres. The edges of the Steiner trees are colored by lines with different colors.
  \item Steiner points were automatically added to each corresponding structure of Steiner tree as shown in black colored spheres of smaller size compared to terminal nodes.
  \item It is possible to obtain compact tree structures despite the presence of non-convex obstacles in the map.
  \item The constructed Steiner trees are able to avoid clutter and overlapping with obstacles. This is due to the fact of using a representation of coordinates rendered from the triangulated free navigable space. As such, nodes and edges of the obtained Steiner trees avoid overlap with the obstacles.
  \item Over various tree configurations, it is possible to observe regions that divide the $360^{\circ}$ into three equal angles. Such configurations are reminiscent to the Fermat-based and hexagonal-based formulations of Steiner trees. In this paper, we make no assumption on the resulting geometries of trees. As such, all obtained tree topologies are the result of the hierarchical bundling of shortest routes and the result of searching for Steiner points over a convex search space by Sequential Quadratic Programming (SQP).
  \item The scenario with \(s=5\) modules of Steiner trees renders relatively larger trees that span wide regions in the map. 
  \item The scenario with \(s=30\) modules of Steiner trees renders smaller tree components with 4-5 terminal nodes.
\end{itemize}

The above-mentioned observations show the potential and the feasibility of generating multiple Euclidean Steiner trees in the plane under arbitrary geometry and configuration of terminal nodes and obstacles in the plane.

In order to show the overall efficiency and size characteristics of the rendered Steiner trees, Fig. \ref{per50v2} shows (a) the time used to compute Steiner trees as a function of ratio $\theta$, (b) the average length of trees of Steiner, and (c) the average number of nodes per Steiner tree. By observing the results from Fig. \ref{per50v2}, one can observe the following facts:

\begin{itemize}
  \item Generating fewer modules of Steiner trees ($\theta=0.05$, $s = 5$) implies rendering Steiner tree structures with more nodes; thus, the higher number of nodes per Steiner tree.
  \item For the above-mentioned cases, the time used to compute each Steiner tree is between 20 and 50s.; whereas trees with a smaller number of nodes can be computed in the order of fewer than 5 s. For instance, computing Steiner trees with \(15-20\) nodes take about 5 s. The large time used in computing relatively larger Steiner trees is due to the shortest path computation in large areas and complex configurations of the visibility graph while taking into account a larger set of terminal nodes.
  \item Generating Steiner trees with $\theta \geq 0.25$ takes about 5 s., and the time to generate Steiner trees remains within the upper bound of 5 s. for large $\theta$.
\end{itemize}

\begin{figure}[t]
\centering
\includegraphics[width=0.98\linewidth]{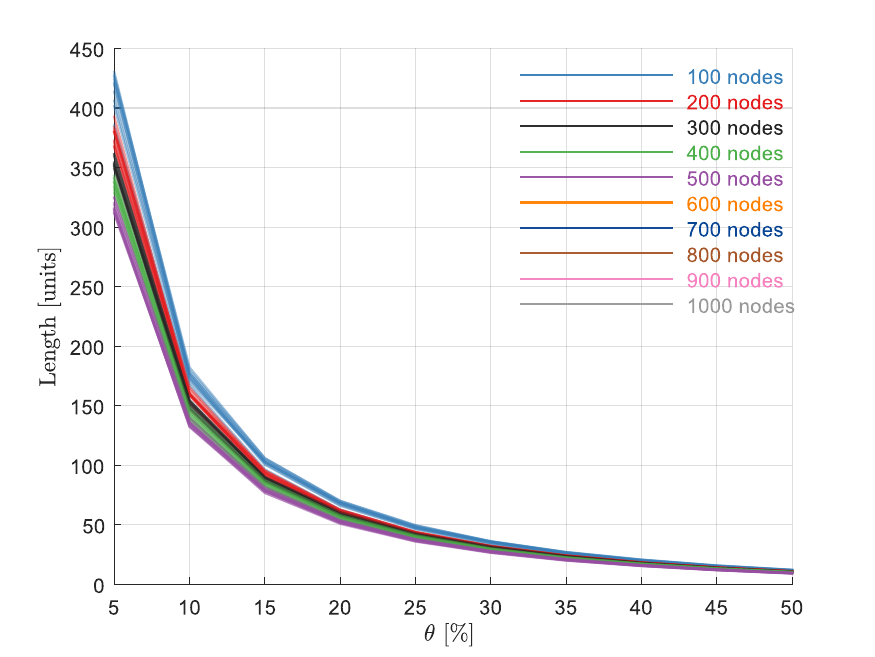}
\caption{Steiner tree length when varying the number of nodes in the plane.}
\label{perv4}
\end{figure}

\begin{figure*}[t!]
\centering
\begin{tblr}{
    colspec = {Q[c]},
    rowsep = 5pt,
    colsep = 1pt,
    vlines = dashed,
    hlines = dashed,
    }
    \stackon{\includegraphics[width=0.19\linewidth]{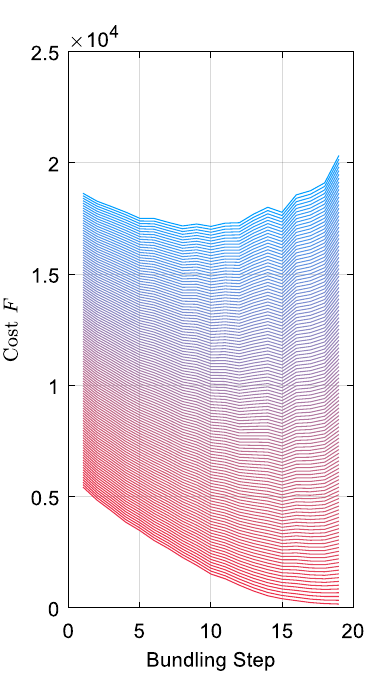}}{\scriptsize Instance 1}
    \stackon{\includegraphics[width=0.19\linewidth]{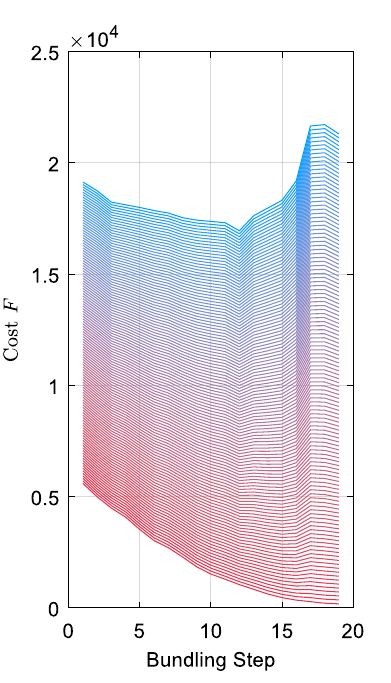}}{\scriptsize Instance 2}
    \stackon{\includegraphics[width=0.19\linewidth]{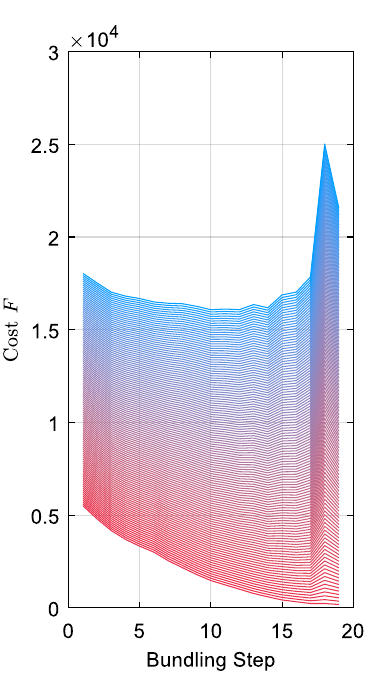}}{\scriptsize Instance 3}
    \stackon{\includegraphics[width=0.19\linewidth]{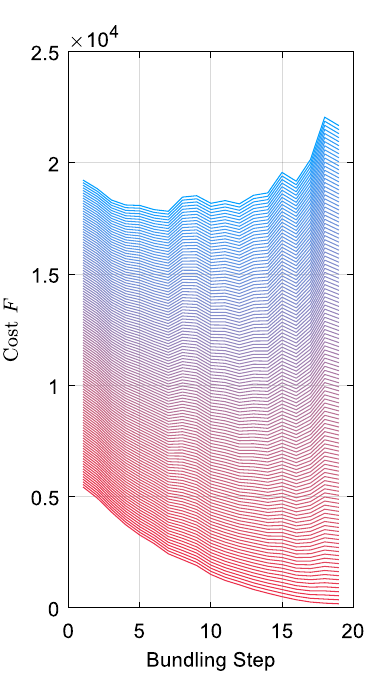}}{\scriptsize Instance 4}
    \stackon{\includegraphics[width=0.19\linewidth]{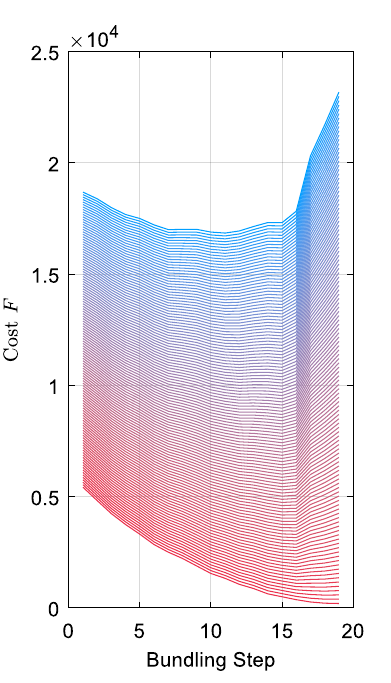}}{\scriptsize Instance 5}
	\\
    \stackon{\includegraphics[width=0.19\linewidth]{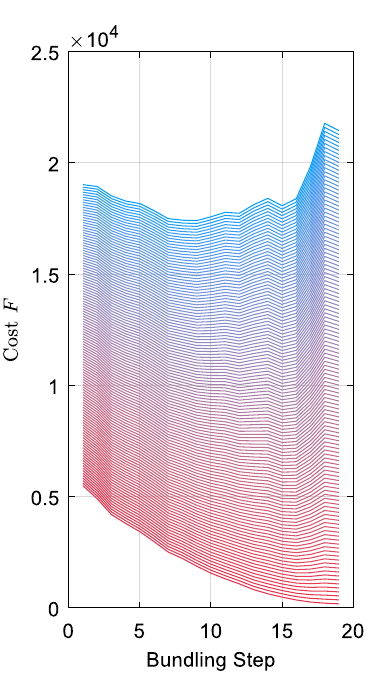}}{\scriptsize Instance 6}
    \stackon{\includegraphics[width=0.19\linewidth]{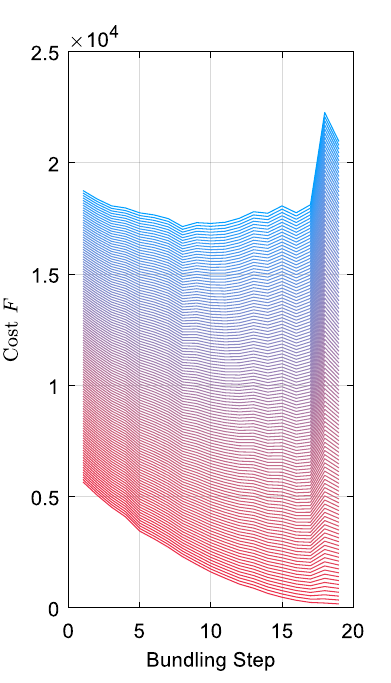}}{\scriptsize Instance 7}
    \stackon{\includegraphics[width=0.19\linewidth]{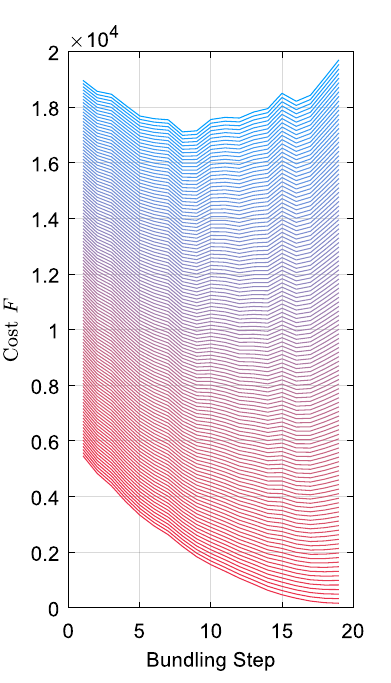}}{\scriptsize Instance 8}
    \stackon{\includegraphics[width=0.19\linewidth]{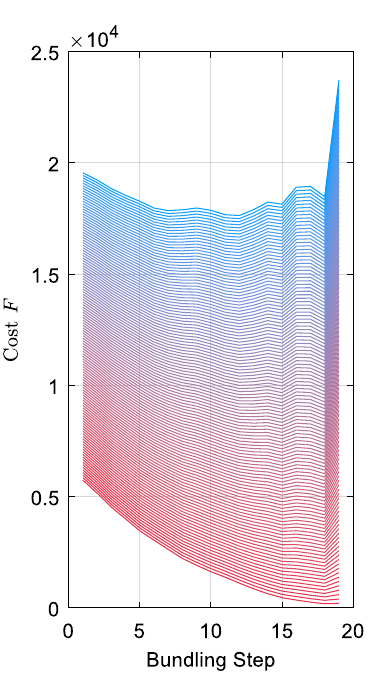}}{\scriptsize Instance 9}
    \stackon{\includegraphics[width=0.19\linewidth]{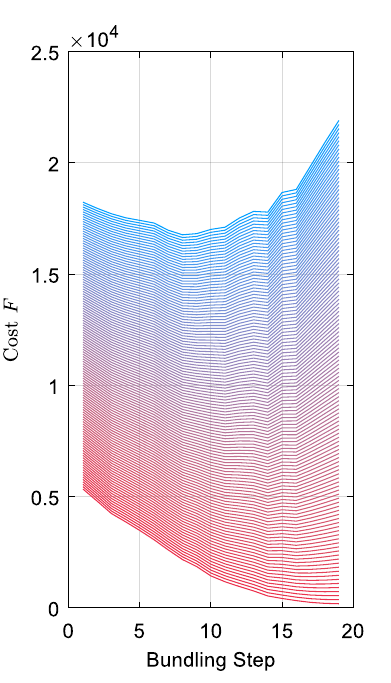}}{\scriptsize Instance 10}
    \\
\end{tblr}
\caption{Instances of the landscape of the cost function, at Eq. (10), for concatenating Steiner trees in the plane considering $w_d = 0.5$ and $w_l$ within the range of $0.1$ (blue color) and $12$ (red color).}
\label{bcosti}
\end{figure*}

In a second set of experiments, we evaluated the overall efficiency when varying the number of nodes in the plane in the range of 100-1000 nodes, and when varying the number of obstacles in the plane. For faster computations, we used an Intel i9-9900K. Fig. \ref{perv3} shows the time to compute Steiner trees as a function of $\theta$. The y-axis of Fig. \ref{perv3} shows the time profiles used to compute Steiner trees for various configurations of the number of nodes in the map (within the range 100-1000 nodes). Fig. \ref{perv3} shows the length of the Steiner trees for larger number of nodes. By observing Fig. \ref{perv3} and Fig. \ref{perv4}, one can note that computing the largest number of Steiner trees takes about 15 - 20 s. and that the overall behaviour of the time performance is consistent with the results obtained in Fig. \ref{per50v2}.

Both Fig. \ref{per50v2} and Fig. \ref{perv3} indicate that to compute larger Steiner trees, it is desirable to generate small-scale Steiner tree components, and then concatenate (bundle) the existing components. The computation of multi-Steiner trees as well as the concatenation (bundling) are prone to parallelization schemes. Studying the parallelization performance and the overall scalability in larger networks is left in the scope of our future work.

In order to show the characteristics of the cost function for Steiner tree concatenation, Fig. \ref{bcosti} shows the landscape of Eq. (\ref{equser}) for ten arbitrary Steiner tree configurations as a function of concatenation step (bundling step). Here, for simplicity and without loss of generality, we used coefficients $w_d = 0.5$, and $w_l$ in the range of $[0.1, 12]$. By observing Fig. \ref{bcosti}, one can note the following facts:

\begin{itemize}
  \item The larger values of $w_l$, as shown by red color in Fig. \ref{bcosti}, penalizes the Steiner tree length; thus the cost function shows a decreasing behaviour suggesting that tree concatenation steps (bundling steps) are necessary to minimize the overall cost function. 
  \item As a result of executing larger number of concatenation steps (bundling steps), fewer modules of Steiner trees can be obtained.
  \item On the other hand, relatively smaller values of $w_l$, as shown by blue-colored lines in Fig. \ref{bcosti}, suggests the existence of u-shaped regions portraying the desirable circumstances to stop tree concatenation.
  \item It is possible to model user-defined cost scenarios for Steiner tree construction and evaluation by using the proposed criterion in Eq. \ref{equser}.
\end{itemize}

The above-mentioned observations on the cost function are essential to allow users to evaluate tradeoffs between the length of the Steiner trees and the installation of potential multiple roots (hubs) in the plane. Furthermore, the above-mentioned results portray the potential to render obstacle-avoiding multi-Euclidean Steiner trees in the plane. It has been observed that such trees are compact, avoid clutter, and overlap with obstacles. Since the search problem of Steiner trees is an NP-hard combinatorial problem, the use of decentralized networks brings an implicit parallelization feature, which has the potential to compute Steiner trees in decentralized regions of the map with utmost efficiency. Also, we used hierarchical clustering as the main grouping method for modules and shortest paths. Investigating the role of clustering and agglomeration methods on Steiner tree geometry has the potential to elucidate tailored operators for enhanced performance.

\section{Conclusion}

In this paper, we explored the feasibility of using a hierarchical bundling approach for constructing obstacle-avoiding multi-Euclidean Steiner trees in the plane. Our method utilizes hierarchical trees to guide the construction of multi-Steiner trees via optimization-based bundling and concatenation of shortest routes and Steiner trees, effectively avoiding clutter and overlap with arbitrary obstacles. This approach not only generates a collection of obstacle-avoiding multi-Euclidean Steiner trees but also concatenates them through hierarchical bundling operations based on a user-defined cost function. Our computational experiments, which involved generating over 500 Steiner trees across 50 maps with various obstacle configurations, demonstrate the feasibility and performance landscape when computing minimal obstacle-avoiding trees in the plane. Investigating the role of additional tree operators in large maps and evaluating decentralized forms of agglomeration strategies is left for future work in our agenda. Also, this paper assumed the fixed roots; investigating the geometry and the optimality of root placement with nonlinear optimization heuristics has the potential to render the versatile instances of multi-Euclidean Steiner. Furthermore, our results have the potential to elucidate new nature-inspired strategies to compute Euclidean Steiner trees in the plane.

\bibliographystyle{IEEEtran}
\bibliography{mybiblio}

\end{document}